\def\year{2021}\relax
\documentclass[letterpaper]{article}
\usepackage{aaai21} 
\usepackage{times} 
\usepackage{helvet} 
\usepackage{courier} 
\usepackage[hyphens]{url} 
\usepackage{graphicx} 
\usepackage{graphicx}  
\usepackage{natbib}
\usepackage{caption}
\usepackage{amsmath}
\usepackage{amsfonts}
\usepackage{bm}
\usepackage{mathtools}
\usepackage{booktabs}
\usepackage{dcolumn}
\newcolumntype{L}{D{,}{\,\pm\,}{-1}}

\DeclareMathOperator*{\argmin}{arg\ min}

\DeclarePairedDelimiter\floor{\lfloor}{\rfloor}

\title{Bayes DistNet - A Robust Neural Network for \\ Algorithm Runtime Distribution Predictions} 
\author{
    Jake Tuero, 
    Michael Buro
    \\
}
\affiliations{
    Department of Computing Science, University of Alberta\\
    Edmonton, Canada\\
    \{tuero, mburo\}@ualberta.ca
}

\begin{document}

\maketitle

\begin{abstract}
    Randomized algorithms are used in many state-of-the-art solvers for constraint
    satisfaction problems (CSP) and Boolean satisfiability (SAT) problems.  For
    many of these problems, there is no single solver which will dominate
    others. Having access to the underlying runtime distributions (RTD) of these
    solvers can allow for better use of algorithm
    selection, algorithm portfolios, and restart strategies.  Previous
    state-of-the-art methods directly try to predict a fixed parametric
    distribution that the input instance follows.  In this paper, we extend RTD
    prediction models into the Bayesian setting for the first time. This new model
    achieves robust predictive performance
    in the low observation setting,
    as well as handling censored observations.
    This technique also allows for richer representations which cannot be achieved
    by the classical models which restrict their output representations. 
    Our model outperforms the previous state-of-the-art model in
    settings in which data is scarce, and can make use of censored data such as
    lower bound time estimates, where that type of data would otherwise be
    discarded. It can also quantify its uncertainty in
    its predictions, allowing for algorithm portfolio models to make better
    informed decisions about which algorithm to run on a
    particular instance.
\end{abstract}

\section{Introduction}

Many of the algorithmic solvers for NP-complete problems, such as CSP and SAT,
rely on backtrack methods. These solvers can have runtimes which vary
substantially, depending on whether they make
mistakes caused by sub-optimal heuristics during the
recursive backtrack calls.  Adding randomization and restarts into the
backtrack algorithms has been shown to help alleviate some of the heavy-tail
nature that these runtimes exhibit, decreasing the algorithm's runtime by many
orders of magnitude in some
cases \cite{Harvey1995NonsystematicBS,10.5555/295240.295710}.

If the underlying \textit{runtime distribution} (RTD) is known for a
particular algorithm, then an optimal fixed cutoff-time restart strategy can
be formulated \cite{Luby1993}. Knowing the underlying RTD can also lead to
efficient use of algorithm portfolios, which can dynamically assign algorithms
to input instances based on the predictive RTD.  These are just a few reasons
why it's important to have models which are robust and can give accurate RTD
predictions for unseen instances.

The current state-of-the-art RTD prediction models are designed to give exact
parametric distributions as output, with the model predicting the particular
distribution parameters \cite{eggensperger2017neural}.
However, these models make
the assumption that the particular algorithms on the given instances follow a
particular parametric distribution.  These models also tend to have their
predictive power hindered in the presence of a small number of observations
per input instance or with censored examples
(only a lower bound for the runtime is known). 
The goal of our research is to
have a robust model under these conditions, and to try to lift some of the
restrictions we put on the models.

The rest of this paper is organized as follows. We first give a summary of the
two areas our research joins for the first time: randomized algorithm runtime
prediction and Bayesian deep learning. We then show how our model extends
existing methods into the Bayesian setting, and show how effective our model
is in several scenarios. Finally, we give some insights into where we think
the next research steps are for extending our work.

\section{Background and Related Work}

In this section, we give an overview of randomized algorithm runtime
prediction and Bayesian learning in neural networks.

\subsection{Randomized Algorithm Runtime Prediction}

Problems such as CSP and SAT are NP-complete, meaning that there is
currently no guarantee there is a polynomial time solution technique.
The dominant solver technique for these types of problems is
backtrack search, in which heuristics pick an unbound variable, assign a value
to it, and the search proceeds recursively. If an inconsistency is detected,
the algorithm backtracks and tries another assignment to that variable.

A consequence of using backtrack-based solvers is that early mistakes can
cause long searches down branches of the search tree which eventually need to
be backtracked. This leads to the so-called ``heavy-tail'' phenomenon, in
which the runtime of these algorithms exhibit tails which are not
exponentially bounded. \citeauthor{Harvey1995NonsystematicBS}
\shortcite{Harvey1995NonsystematicBS} was the first to show that adding
randomization and restarts into the backtracking algorithm can alleviate the
issue of wasting time on eventual dead-ends.
\citeauthor{10.5555/295240.295710} \shortcite{10.5555/295240.295710} presented
general methods for introducing controlled randomization, and showed that
speedups of several orders of magnitude could be achieved for state-of-the-art
algorithms.

Introducing randomness into algorithms means that a given algorithm run on the
same input problem instance has a runtime which follows some
initially unknown distribution. There are
many reasons why one would want the ability to predict such
runtime distributions (RTDs).
\citeauthor{Luby1993} \shortcite{Luby1993} showed that if
one knows the underlying RTD, then it's
possible to construct a fixed cutoff-time restart strategy that is optimal
among all possible universal strategies, up to a constant factor.
Other popular CSP and SAT solvers use a portfolio of various
algorithms, such as SATzilla \cite{xu2008satzilla} and ArgoSmArT
\cite{10.1007/978-3-642-02777-2_31}. Knowing the RTDs of each algorithm in the
portfolio allows such solvers to choose the \textit{best} algorithm for a
given problem instance.

The majority of the early work for predicting algorithm runtimes involved
predicting mean instance runtimes, given the instance's features.  Many of
these methods were based on regression variants, such as ridge regression used
in SATzilla \cite{xu2008satzilla}.

\citeauthor{gagliolo2005neural}
\shortcite{gagliolo2005neural} investigated using a neural network to predict
the time remaining before an algorithm reaches the solution on a given problem
instance, in the context of algorithm portfolio time allocation.
Previous methods for predicting RTDs had separate models for each distribution
parameter.  DistNet \cite{eggensperger2017neural} established a new
state-of-the-art RTD prediction model which jointly learns the parameters of
the RTD by using a neural network.  Each output node of the neural network
corresponds to a parameter for the given distribution.  The neural network's
loss function directly minimizes the negative log-likelihood (NLLH) of the
distribution parameters, given the observed runtimes.

\subsection{Bayesian Neural Networks}

Traditional neural networks can be viewed as a probabilistic model: given a
dataset $\mathcal{D} = \{(x_i, y_i)\}_{i=1}^N$, the parameterized model
$P(\mathbf{y} | \mathbf{x},\mathbf{w})$ takes in a $d$-dimensional input
$\mathbf{x} \in \mathbb{R}^d$ and computes a point estimate for each output
$\mathbf{y} \in \mathcal{Y}$.  Training most commonly involves finding the
set of weights $\mathbf{w}$ which
maximizes the likelihood of the data.

These networks can be prone to overfitting, especially when the number of 
model parameters is sufficiently greater than the number of training samples. 
Regularization techniques like using weight priors and 
dropout \cite{srivastava2014dropout} have been proposed to mitigate overfitting. 

Bayesian neural networks (BNNs) \cite{mackay1992practical,neal2012bayesian}
extended traditional neural networks by replacing the network's deterministic
weights with a prior distribution over these weights, and using
posterior inference. The full posterior distribution of the model's
parameters $\mathbf{w}$ is used when making predictions of unseen
data. Prediction for this separate probabilistic model involves taking the
expectation over the optimized posterior distribution $P(\mathbf{w} |
\mathcal{D})$. The \textit{posterior predictive distribution} of unseen data
$(\hat{\mathbf{x}}, \hat{\mathbf{y}})$ is given by
\begin{align*}
P(\hat{\mathbf{y}} | \hat{\mathbf{x}}) &= \mathbb{E}_{P(\mathbf{w} | \mathcal{D})} \left[ P(\hat{\mathbf{y}} | \hat{\mathbf{x}}, \mathbf{w}) \right] \\
&= \int P(\hat{\mathbf{y}} | \hat{\mathbf{x}}, \mathbf{w}) P(\mathbf{w} | \mathcal{D}) \ d \mathbf{w}.
\end{align*}

Bayesian neural networks can help with overfitting as the model implicitly
involves using an ensemble of an infinite number of neural networks by
averaging over all possible weight values, 
while adding a constant multiple number of 
network parameters, determined by
how many parameters the parametric distribution has.
Bayesian neural networks are also able to capture both \textit{aleatoric}
uncertainty and \textit{epistemic} uncertainty in its predictions.
Aleatoric uncertainty captures uncertainties which are inherent to running
statistical trials, like the outcome of a dice toss.  This type of uncertainty
\textit{cannot} be reduced, no matter how much data is collected.  Epistemic
uncertainty captures the uncertainty in the model being used, which is due to
limited data and/or knowledge. This type of uncertainty \textit{can} be
reduced with more data.

Both inference and prediction for BNNs involve calculating the posterior 
\begin{equation*}
P(\mathbf{w} | \mathcal{D}) = \frac{P(\mathcal{D} | \mathbf{w}) P(\mathbf{w})}{P(\mathcal{D})} 
    = \frac{P(\mathcal{D} | \mathbf{w}) 
    P(\mathbf{w})}{\int P(\mathcal{D} | \mathbf{w'}) P(\mathbf{w'}) \ d \mathbf{w'}}.
\end{equation*}
The prior $P(\mathbf{w})$ is
something we can choose, but the integral involved in the posterior
is often computationally intractable. While the above posterior can
be approximated using sampling-based inference algorithms like Markov Chain
Monte Carlo (MCMC) methods \cite{gelfand1990sampling}, these do not scale well
as the number of samples and/or parameters increase
\cite{blei2017variational}, as is the case in neural networks.  An
alternative, suggested by \citeauthor{hinton1993keeping}
\shortcite{hinton1993keeping} and \citeauthor{graves2011practical}
\shortcite{graves2011practical}, is to use a variational approximation for the
posterior $P(\mathbf{w} | \mathcal{D})$.  \textit{Variational methods}
construct a new distribution $q(\mathbf{w} | \bm{\theta})$, parameterized by
$\bm{\theta}$, that approximates the true posterior $P(\mathbf{w} |
\mathcal{D})$ by minimizing the Kullback-Leibler (KL) divergence between the
two:
\begin{align*}
\bm{\theta}_{\text{VI}} &= \argmin_{\bm{\theta}} KL 
    \left[ q(\mathbf{w} | \bm{\theta}) || P(\mathbf{w} | \mathcal{D}) \right] \\
&= \argmin_{\bm{\theta}} \int q(\mathbf{w} | \bm{\theta})
    \log \frac{q(\mathbf{w} | \bm{\theta})}{P(\mathbf{w}) P(\mathcal{D} | \mathbf{w})} \ d \mathbf{w} \\
&= \argmin_{\bm{\theta}} KL 
    \left[ q(\mathbf{w} | \bm{\theta}) || P(\mathbf{w}) \right] - 
    \mathbb{E}_{q(\mathbf{w} | \bm{\theta})} \left[ \log P(\mathcal{D} | \mathbf{w}) \right].
\end{align*}
We denote $\bm{\theta}_{\text{VI}}$ as the parameters found
from variational inference methods.
Bayes by Backprop \cite{blundell15} uses the above to form the
cost function
\begin{multline}\label{bbb1}
J_{\text{BBB}}(\mathcal{D}, \bm{\theta}) = KL \left[ q(\mathbf{w} | \bm{\theta}) \ ||  \ P(\mathbf{w} ) \right] \\
- \mathbb{E}_{q(\mathbf{w} | \bm{\theta})} \left[ \log P(\mathcal{D} | \mathbf{w}) \right].
\end{multline}
The intuition for the above cost function is
that there is a tradeoff between having a variational posterior that is
\textit{close} to the prior we choose
yet also able to explain the likelihood of the
data. Despite the simple formulation of the above, the cost
$J_{\text{BBB}}(\mathcal{D}, \bm{\theta})$ is still intractable as it
involves calculating posteriors over high-dimensional spaces. While
neural networks would seem like an appropriate tool to find a function which
minimizes $J_{\text{BBB}}$, we cannot naively apply backpropagation through
nodes which involve stochastity. 
By applying a generalization of the re-parameterization trick
\cite{opper2009variational,Kingma2014AutoEncodingVB},
Bayes by Backprop \cite{blundell15} then provides
an approximation to the cost (Eq.~\ref{bbb1}) as
\begin{multline}\label{bbb_cost_1}
J_{\text{BBB}}(\mathcal{D}, \bm{\theta}) \approx \sum_{i=1}^n \log q(\mathbf{w}^{(i)} | \bm{\theta}) - \log P(\mathbf{w}^{(i)}) \\
- \log P(\mathcal{D} | \mathbf{w}^{(i)}),
\end{multline}
where $\mathbf{w}^{(i)}$ is the set of weights drawn from the $i$-th Monte
Carlo sample of the variational posterior $q(\mathbf{w} |
\bm{\theta})$.
Having a computational efficient way of training BNNs has
allowed extensions to other network architectures like RNNs
\cite{fortunato2017bayesian} and CNNs \cite{shridhar2019comprehensive}, as
well as studies into utilizing the uncertainty measures the BNNs provide
\cite{amodei2016concrete,kendall2017uncertainties}.


\section{Problem Formulation}

We follow the same problem statement introduced by 
\citeauthor{eggensperger2017neural} \shortcite{eggensperger2017neural}: A
randomized algorithm $A$ is run on a set of $n$ problem instances
$\Xi_{\text{train}} = \{\xi_1, \dots, \xi_n\}$. Each instance $\xi \in
\Xi_\text{train}$ has $m$ instance features $\mathbf{f}(\xi) = [f(\xi)_1,
\dots, f(\xi)_m]$. Since algorithm $A$ is randomized, $k$ runtime
observations $\mathbf{t}(\xi) = [t(\xi)_1, \dots, t(\xi)_k]$ are gathered by
executing $A$ on problem instance $\xi$, with $k$ different random
number generator seeds. The goal is to learn a model that can predict the
RTD for unseen instance $\xi_{n+1}$, given features $\mathbf{f}(\xi_{n+1})$.

Because it is common during the data generation stage to terminate
the algorithm on \textit{hard} problem instances which take a long time to
solve, the exact runtime may not be known.
Ideally, we would like to make use of the
measured runtime lower bound without discarding the data and wasting
time.
From this motivation, we also consider cases where there is
censoring, i.e., stopping the algorithm execution if it exceeds a cutoff time
$t_c$.  We define the level of censoring as the fraction
of runtime observations which are censored
due to exceeding a cutoff time $t_c$. If there are $N$
runtimes in total, with a predetermined level of
censoring of $c$, then $t_c$ is defined as the $u$-th fastest runtime where $u
=\floor*{N(1-c)}$.
The adjusted runtimes $t'(\xi)_i$ are thus set
as $\text{min}(t(\xi)_i, t_c)$ for all $i$.

\section{A Bayesian Approach to Predicting RTDs}

Previous methods for predicting RTDs would fit RTD parameters
to the set of runtimes for
each training instance,
then measure loss in the space of
RTD parameters $\bm{\beta}$.  \citeauthor{eggensperger2017neural}
\shortcite{eggensperger2017neural} introduced DistNet, a new
state-of-the-art algorithm runtime prediction model, which is the first of
its kind that jointly learns all RTD parameters (as opposed to having
independent models for each RTD parameter) by directly minimizing the NLLH
loss function.

In this section, we extend DistNet to the Bayesian setting 
to obtain a more robust model in both low sample settings, and for handling
censored observations.

\subsection{DistNet - A State-of-the-Art Algorithm RTD Prediction Model}

Before we introduce our extensions, we give an overview of the DistNet model
\cite{eggensperger2017neural}.  DistNet is a simple feed-forward network for a
given distribution $\mathcal{F}$ with distribution parameter vector
$\bm{\beta}$.  There is one input neuron for each instance feature, and one
output neuron for each distribution parameter $\beta_i$. 
The novelty of DistNet is
that it directly minimizes the NLLH of the chosen
distribution $\mathcal{F}$. The loss function used in DistNet is
\begin{equation}\label{distnet_loss_func}
	J_{\text{DN}}(\mathbf{w}) = - \sum_{\xi \in \Xi_{\text{train}}} \sum_{i=1}^k
        \log \mathcal{L}_\mathcal{F} (\hat{\bm{\beta}}_{\mathbf{w}, f(\xi)} |
        t(\xi)_i),
\end{equation}
where $\mathcal{L}_\mathcal{F}$ is the
likelihood function of the parametric distribution $\mathcal{F}$, and
$\hat{\bm{\beta}}_{\mathbf{w}, f(\xi)}$ are the
parameters of the distribution from the network output, with current weights
$\mathbf{w}$.

\subsection{Extending DistNet with Bayesian Inference}

To extend DistNet with Bayesian inference, which we will refer to as
{\em Bayes DistNet}, a few changes need to be made to the network. We
start from the same base network, which has one input neuron for each input
feature, and uses simple feed-forward layers.

While the straightforward choice for the Bayesian network output would be the
parameters $\bm{\beta}$ of the chosen distribution $\mathcal{F}$, the actual
implementation is non-trivial and is computationally expensive.  The network
would produce a posterior over distribution parameters by sampling the network
repeatedly, and these parameters would then be used in the likelihood
$\mathcal{L}_\mathcal{F} (\hat{\bm{\beta}}_{\mathbf{w}, f(\xi)} | t(\xi)_i)$, with the
likelihood weighted by the posterior, resulting in a posterior predictive distribution.  
However, this requires a density estimation
method using the set of network samples $\hat{\bm{\beta}}_{\mathbf{w}, f(\xi)}$, such
that the gradient information can still be tracked from the loss function.

Instead, we choose to have the Bayesian network directly output predicted
runtimes $\hat{y}$. By sampling the network, a sequence
of runtimes is produced which is assumed to follow $\mathcal{F}$. The
parameters of $\mathcal{F}$ can then be approximated by computing their
maximum likelihood estimates (MLEs), denoted
$\hat{\bm{\beta}}^{\text{MLE}}$. The choice of
$\mathcal{F}$ can ensure that we have analytical formulations of
$\hat{\bm{\beta}}^{\text{MLE}}$, which are fast to compute
using deep-learning framework intrinsics, and whose gradients can be tracked
from the loss function.

To make the network Bayesian, we need to define a
variational posterior $q(\mathbf{w} | \bm{\theta})$,
the prior of the network weights $P(\mathbf{w})$,
and the likelihood of the training data.
Following \citeauthor{blundell15} \shortcite{blundell15}, we
assume that the variational posterior $q(\mathbf{w} | \bm{\theta})$
is a diagonal Gaussian distribution with mean $\bm{\mu}$ and standard
deviation $\bm{\sigma}$. 
While it is possible to use a multivariate Gaussian,
a diagonal Guassian allows for simple computations without major numerical issues.
To be able to use backpropagation, we use
the re-parameterization given by \citeauthor{blundell15}
\shortcite{blundell15}, which obtains a sample of weights $\mathbf{w}$ for
each layer by the following procedure: Sample the parameter-free noise
$\bm{\epsilon} \sim \mathcal{N}(\bm{0},\mathbf{I})$, and let $\bm{\sigma} =
\log(\bm{1}+\exp(\bm{\rho}))$. We can then denote $\bm{\theta} = (\bm{\mu},
\bm{\rho})$, and the parameterization of $\bm{\sigma}$ ensures that the values
are always $>0$. The weights $\mathbf{w}$ can then
be sampled by $\mathbf{w} = \bm{\mu} + \bm{\sigma} \circ \bm{\epsilon}$, where
$\circ$ is point-wise multiplication. Once a particular set of point-wise
weights is sampled for each of the layers of the network, those weights are
then used as in the traditional feed-forward layers.

Since the network's task is to predict runtimes directly, it has a single
output neuron. The network output $\hat{y}$ represents the predicted runtime
for a particular set of point-wise weights which were sampled from each
layer. To get a predictive distribution, Monte Carlo sampling is used, in
which the same input to the network will be used $N$ times to produce $N$
separate output values.

From Eq.~\ref{bbb_cost_1}, the term $\log P(\mathcal{D} | \mathbf{w})$
is replaced with the likelihood from $J_{\text{DN}}$,
which measures the likelihood of the data in our new model.
The variational approximated cost function for the network now becomes
\begin{multline}\label{bbb_cost_2}
    J(\mathcal{D}, \bm{\theta})_{\text{BDN}} = \sum_{i=1}^N \log q(\mathbf{w}^{(i)} |
    \bm{\theta}) - \log P(\mathbf{w}^{(i)}) \\ 
    - \sum_{\xi \in \Xi_{\text{train}}} \sum_{j=1}^k
    \log \mathcal{L}_\mathcal{F} \big(\hat{\bm{\beta}}^{\text{MLE}} |
    t(\xi)_j \big),
\end{multline}
where $\hat{\bm{\beta}}^{\text{MLE}}$ is $\mathcal{F}$'s distribution
parameter vector computed by MLE using the $N$ predicted runtimes $\hat{y}_i$,
and each $\hat{y}_i$ is produced by the network using weights
$\mathbf{w}^{(i)}$ from the $i$-th Monte Carlo sample.

\subsection{Observing Censored Runtimes}

In our application domain, censoring occurs when an
algorithm needs to be stopped before completion, i.e., we only know a lower
bound on the time it takes to actually complete the algorithm.  There is a
rich history in survival analysis on handling different types of censored
data, and we recommend \citeauthor{klein03} \shortcite{klein03} for background
reading.  \citeauthor{Gagliolo2006} \shortcite{Gagliolo2006} first introduced
handling \textit{Type I censored sampling} in the context of algorithm runtime
prediction, and we give the explanation for clarity.

If censoring time of $t_c$ is used, then $t_c$ is the maximum time an
algorithm can run before being terminated. The dataset 
$\mathcal{D} = \{(\bm{f}(\xi)_i,
t(\xi)_i)\}_{i=1}^N$ from before now becomes 
$\mathcal{D} = \{(\bm{f}(\xi)_i, t_i,
\delta_i)\}_{i=1}^N$, where $t_i = \min(t(\xi)_i, t_c)$, and $\delta_i$ is
a Boolean variable indicating
that an individual sample was not censored (i.e., $\delta_i = 1$ indicates
that $t(\xi)_i \le t_c$).

Let $f_{\mathcal{F}}$ denote the density of distribution $\mathcal{F}$
with parameters $\bm{\beta}$, and let $\mathcal{S}_{\mathcal{F}}$ denote the
survival function of distribution $\mathcal{F}$.
If the runtime is not censored, then we observe $\delta_i = 1$, and its
contribution to the likelihood function is the density function at that time
as it normally would be,
\begin{equation*}\label{censored1}
    \mathcal{L}_\mathcal{F} \big(\hat{\bm{\beta}}^{\text{MLE}} |
    t_j \big) = f_{\mathcal{F}} \big( t_j | 
    \hat{\bm{\beta}}^{\text{MLE}} \big).
\end{equation*}
If the runtime is censored, then we observe $\delta_i = 0$,
and all we can say under censoring is that the runtime exceeds
the cutoff $t_c$.  
This is equivalent to the survival function, and so its contribution
to the likelihood function when censored is
\begin{equation*}\label{censored2}
    \mathcal{L}_\mathcal{F} \big(\hat{\bm{\beta}}^{\text{MLE}} |
    t_j \big) = \mathcal{S}_{\mathcal{F}} \big( t_j | 
    \hat{\bm{\beta}}^{\text{MLE}} \big).
\end{equation*}

Combining the above, we get
\begin{equation}\label{censored3}
    \mathcal{L}_\mathcal{F} \big(\hat{\bm{\beta}}^{\text{MLE}} |
    t_j \big) = f_{\mathcal{F}} \big( t_j | 
    \hat{\bm{\beta}}^{\text{MLE}} \big)^{\delta_j}
    \mathcal{S}_{\mathcal{F}} \big( t_j | 
    \hat{\bm{\beta}}^{\text{MLE}} \big)^{1-\delta_j},
\end{equation}
which is then used to replace the likelihood term in the
Bayesian cost function (Eq.~\ref{bbb_cost_2}) from above.

\subsection{Applying Bayes DistNet to RTD Prediction}

For comparative results, we follow the same preprocessing steps from
\citeauthor{eggensperger2017neural} \shortcite{eggensperger2017neural}. Input
instance features are standardized to mean 0 and standard deviation 1, and
observed runtimes are scaled into the range of $[0,1]$. For a level of
censoring $c$, where $c$ is the fraction of
censored runtimes from a total of $N$, the cutoff time $t_c$ is the $u$-th
fastest runtime, where $u = \floor*{N(1-c)}$.  The runtimes are then set to
$\min(t(\xi)_i, t_c)$, and the observations include whether censoring occurred
or not.

Both the reference DistNet model and our new Bayesian variant share the
majority of the network architecture proposed by \citeauthor{eggensperger2017neural}. 
The input layer has a neuron for each
instance feature. This is then followed by two hidden fully connected
layers, each of which has 16 neurons. DistNet has a fully
connected output layer with one neuron per parameter for the specific
parametric distribution, whereas Bayes DistNet has a single output.

The Bayes DistNet architecture also has a prior $P(\mathbf{w})$ and a
variational posterior $q(\mathbf{w} | \bm{\theta})$ that needs to be
specified.  As previously noted, we use a diagonal Gaussian variational
posterior, with initial parameters $\bm{\theta}_\text{init} =
(\bm{\mu}_\text{init}, \bm{\rho}_\text{init})$, $\bm{\mu}_\text{init} \sim
\mathcal{N}(0, 0.1)$ and $\bm{\rho}_\text{init} \sim \mathcal{N}(-3, 0.1)$.
The chosen $\bm{\mu}_\text{init}$ ensures weights are initialized around 0
(just as we would for a standard neural network), and $\bm{\rho}_\text{init}$
is initialized to a small value, as we found learning is not as stable
otherwise \cite{shridhar2019comprehensive}.
For the prior, we follow the proposal
from \citeauthor{blundell15} \shortcite{blundell15} of using
a mixture of two Gaussian distributions
\begin{equation*}
  P(\mathbf{w}) = \prod_i \alpha \mathcal{N} (\mathbf{w}_i; 0, \sigma_1^2) +
  (1-\alpha) \mathcal{N} (\mathbf{w}_i; 0, \sigma_2^2),
\end{equation*}
where $\mathcal{N}(x; \mu, \sigma^2)$ is the Gaussian density 
evaluated at $x$ with mean $\mu$ and variance $\sigma^2$, with
$\alpha=0.5$, $\sigma_1 = 0.3$, and $\sigma_2 = 0.01$.
The choice of these parameters did not have a noticeable difference
so long as $\sigma_1 > \sigma_2$ and $\sigma_2 \ll 1$,
as per \citeauthor{blundell15} \shortcite{blundell15}.
When performing
inference and prediction, we use 16 Monte Carlo samples. As with previous studies
\cite{fortunato2017bayesian,shridhar2019comprehensive}, a large number of
samples gives marginally better results at the cost of computation time. We
did not find much improvements beyond 16 samples.

DistNet uses the \textit{tanh} activation function,
with the exception of the \textit{exponential} activation 
function on the output layer
to ensure the output distribution parameters are $>0$.
Bayes DistNet, on the other hand,
uses \textit{softplus} \cite{glorot2011deep} on all layers which is a smooth
approximation to the recitifer activation function, as we want to ensure
that activations for variational posterior variances never become $\le 0$.
We then largely follow the
hyperparameter choices from \citeauthor{eggensperger2017neural}
\shortcite{eggensperger2017neural}, whose DistNet model we are basing our Bayes
DistNet model from.  Both network architectures use stochastic gradient
descent (SGD), with batch normalization \cite{ioffe2015batch},
L$_2$-regularization of $1e^{-4}$, a learning rate of $1e^{-3}$ which
exponentially decays to $1e^{-5}$ over 500 expected epochs, and gradient
clipping of $1e^{-2}$.  Early stoppage \cite{prechelt1998automatic} is used on
a separate validation set to reduce overfitting.


\section{Experiments}

The focus of our experiments is to see how our Bayesian model compares with
the current state-of-the-art in low observation number settings, and with
various levels of censorship. When gathering
data to train the models, using a lower number of observations allows for less
time required for the data gathering process. Moreover, being able to handle
censored examples also means that
one can still use instances which stop prematurely as training data,
instead of throwing them away.
There are also many domains in which much of the data is censored, such as
survival data in medical trials, or nodes that remain
in the open list when heuristic search methods are used.

For both experiments, we look at two different algorithms run on different
problem instances, with the data gathered by
\citeauthor{eggensperger2017neural} \shortcite{eggensperger2017neural}:
\begin{itemize}
    \item \textbf{Clasp-factoring}: The \textit{Clasp} \cite{Gebser2012} CDCL
      solver running on SAT-encoded factorization problems.
    \item \textbf{LPG-Zenotravel}: The \textit{LPG} \cite{Gerevini2002} local
      search solver for planning graphs running on the \textit{Zenotravel}
      planning domain \cite{Penberthy1994}.
\end{itemize} 
Both DistNet and Bayes DistNet require a parametric
distribution: DistNet directly outputs parameters for this distribution, and
both use the distribution in the loss function.  For both of the above
scenarios, \citeauthor{eggensperger2017neural} \shortcite{eggensperger2017neural}
considered different parametric distributions to use for DistNet.  By using
the Kolmogorov-Smirnov (KS) goodness-of-fit test, they chose the top two
distributions to use in their comparison.  For both \textit{Clasp-factoring}
and \textit{LPG-Zenotravel}, the inverse Gaussian and lognormal distribution
were found to have the closest fit to the empirical distributions as compared
to the other considered distributions.  In our experiments, we also use these
two distributions.  The training process uses Python PyTorch 1.5
\cite{NEURIPS2019_9015}, and we make use of an open source Bayesian Layers
PyTorch library \cite{shridhar2019comprehensive}. 
All experiments were run on an Intel i7-7820X and Nvidia GTX 1080 Ti,
with 64GB of memory running Ubuntu 16.04.

\subsection{Low Sample Count per Instance}

\begin{figure*}[t]
  \centering
  \includegraphics[width=0.88\textwidth]{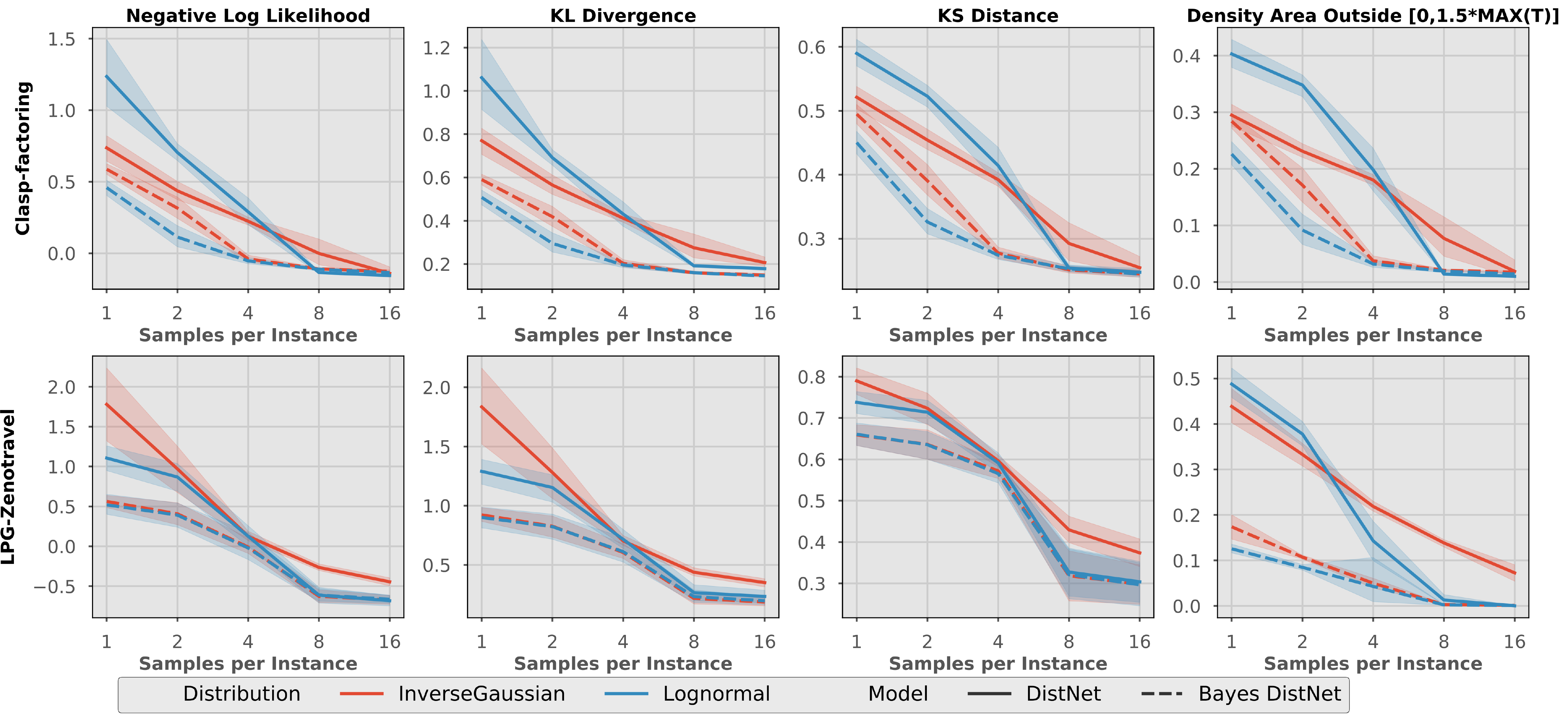}
  \caption{Comparison of DistNet and Bayes DistNet, for various
    numbers of observed runtimes per instance. Metrics are averaged across
    10-folds, repeated for multiple seeds.  From left to right: NLLH,
    KL-Divergence, KS-Distance, and percentage of the density area outside the
    expected range of 0 to 1.5 times the maximum observed runtime for each
    instance.}
\label{fig_low_samples}
\end{figure*}

We evaluated the performance of our Bayes DistNet model by varying the
number of observed runtimes per instance.  Every instance starts with 100
observed runtimes, and a random subset of those 100 are selected.  Having a
smaller number of observed runtimes per instance means that the total training
set is reduced.  For testing, we keep all 100 observed runtimes for every
instance in the test set.  The evaluation is performed using a 10-fold
cross-validation, repeated with multiple seeds, and
aggregating the results.

Figure \ref{fig_low_samples} reports several metrics to evaluate the
goodness-of-fit of each model, compared to the empirical observed runtimes of
the test set.  The likelihood is a way to measure how probable a model is,
given the observed data.  A lower negative log likelihood (NLLH) thus
indicates that one particular model gives the observed data a higher
probability of occurring.  For both scenarios of \textit{Clasp-factoring} and
\textit{LPG-Zenotravel}, Bayes DistNet achieves a lower negative log
likelihood across various levels of samples per instance.  Given enough data,
both DistNet and Bayes DistNet converge to having an equal
likelihood score.

While the likelihood is a way to measure how probable a model is given the
data, it does not indicate the shape of the predicted runtime
distribution. We thus looked at two metrics to quantity how dissimilar the 
model's predictive distribution shape is compared to the empirical
distribution, namely KL-Divergence and the Kolmogorov-Smirnov statistic
(KS-Distance).  The KL-Divergence quantifies the amount of information
\textit{lost} if the model's distribution is used instead of the empirical
distribution.  The KS-Distance measures the maximal distance in height from
the model's CDF to the empirical CDF. For both metrics, lower is better.

Our Bayesian model achieves lower KL-Divergence and lower KS-Distance with
respect to the empirical distribution, as compared to the DistNet model across
various levels of samples per instance.  This indicates that our Bayesian
model is able to produce predictive RTDs which are closer in shape to the
empirical RTD.  Again, given enough data, both DistNet and our Bayesian model
converge to predicting RTDs which on average are equally close to the
empirical RTDs.

\begin{figure*}[t]
    \centering
    \includegraphics[width=0.88\textwidth]{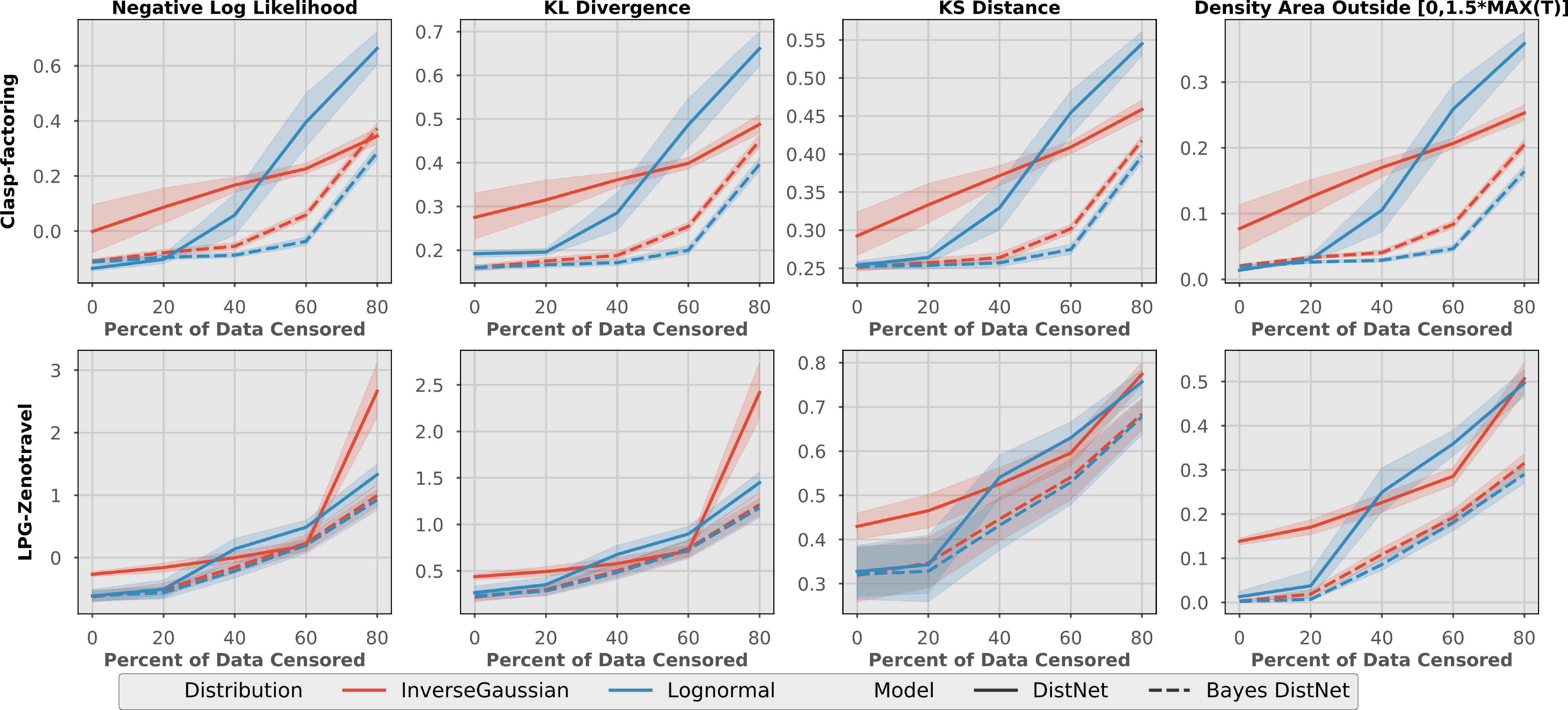}
    \caption{Comparison of DistNet and Bayes DistNet, for various
      levels of censoring, each with 8 observations per instance.  All metrics are
      averaged across 10-folds, repeated for multiple seeds. From left to right:
      NLLH, KL-Divergence, KS-Distance, and percentage of density area outside the
      expected range of 0 to 1.5 times the maximum observed runtime for each
      instance.}
    \label{fig_censoring_samples}
  \end{figure*}

\subsection{Handling Censored Observations}

We also evaluated our Bayes DistNet model by keeping the number of
samples per instance constant, but varying the percentage of censoring in the
training data. We decided to use 8 samples per instance, as this was the
number of samples we start to see both DistNet and our Bayesian variant
perform similarly without censoring. 
To have an accurate comparison, we modified the DistNet loss function
to include censored samples as well, following Eq.~\ref{censored3}.
Figure \ref{fig_censoring_samples}
reports similar metrics as before to evaluate the goodness-of-fit for both
DistNet and our Bayesian variant, as compared to the empirical RTDs.

The NLLH of both the best DistNet model and Bayes DistNet are
identical under no censoring.  As the censoring percentage increases, both
models are trained on less fully observable data, and we expect the NLLH to
increase.  Interestingly, we see that while both models have their NLLH
increase, Bayes DistNet increases at a slower rate than DistNet.
This indicates that our Bayesian model is better able to capture the
uncertainty that comes with using censored data.

Using the distance metrics again, we can see how similar our Bayesian model is
to the empirical RTD, as compared to DistNet.  We expect both the
KL-Divergence and the KS-Distance to increase as the percentage of censored
data increases.  Both the best DistNet model and our Bayesian model start
roughly with similar distance metrics. As the percentage of censored data
increases, our Bayesian model's distance metrics increase at a slower rate
compared to DistNet.  We see that our Bayesian model is able to produce
distributions which are more similar to the empirical RTDs under censoring, as
compared to DistNet.

\subsection{Utilizing Model Uncertainty to Recognize Overconfidence}

\begin{figure}[t]
\centering
\includegraphics[width=0.85\columnwidth]{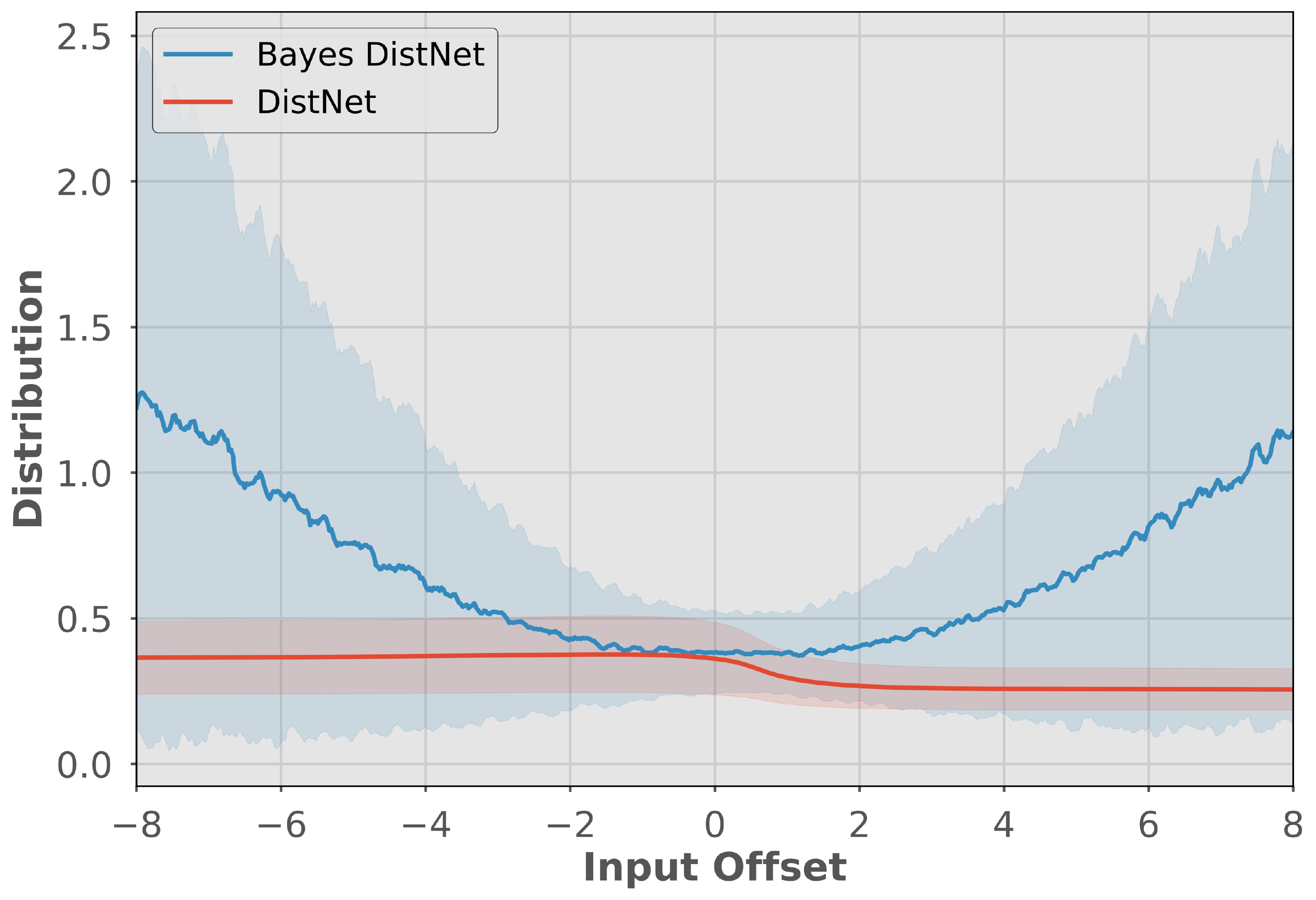}
\caption{Predictive distributions of adversarial inputs for DistNet
  and Bayes DistNet. Both models are trained on the \textit{Clasp-factoring}
  dataset, using the lognormal distribution loss function.
  Solid lines are mean predictions, with shaded regions
  showing interquartile ranges.}
\label{fig_uncertainty}
\end{figure}

A known issue with standard neural networks, which use point estimates for
weights, is that they tend to be overconfident in regions which had little or
no data during training.  Neural networks fit a function to training data
which minimizes the chosen loss function.  If deterministic weights are used,
then a single particular function of many possible ones is chosen for the
data. This function is then extrapolated to make predictions for inputs
which may be \textit{far} from the data used to train the network, resulting
in overconfident predictions. As a result, adversarial and out-of-distribution
inputs can be given to the network, and the network would give predictions
which it is confident in, which can be exploited by attackers.

Instead of using point estimates for weights, Bayesian neural networks make
predictions using all possible weight values, weighted by their posterior
probability.  We can think of this as a type of model averaging.  When
predictions are made for inputs which are far from the data used to
train the network, the model averaging considers that there are many possible
extrapolations, which result in the confidence regions diverging.

Figure~\ref{fig_uncertainty} shows the result of traditional neural networks,
as is the case for the original DistNet model, of being overconfident in its
predictions for adversarial inputs.  
Both networks were trained on the \textit{Clasp-factoring} dataset,
except for a held out sample,
and both models use the lognormal distribution in the loss function.
The held out sample then had all its features shifted by values in the
range [-8,8] to produce a spectrum adversarial samples.

The predictive distribution mean and interquartile ranges were then plotted
for these samples.  DistNet has small interquartile ranges for
adversarial samples, 
which suggests it does not see adversarial samples any different than the ones
it was trained on. 
This can be problematic if
we were to use DistNet in sensitive applications, or scenarios where
computation is expensive.
On the other hand, 
Bayes DistNet has interquartile ranges which diverge on 
adversarial samples.
This suggests that
Bayes DistNet has the capability to indicate when it is not confident in
its predictions. In sensitive or computationaly expensive scenarios, a
fallback measure can then be used when the predictive distribution's
interquartile range falls outside a given threshold.

\subsection{Discussion}

We have shown that in scenarios featuring a low number of instances
or censored observations, our Bayesian model is able to generate
distributions which give higher likelihoods of the data,
and better match the empirical RTD's shape.

Both DistNet and Bayes DistNet require a parametric distribution.
Both use this distribution in the loss function for the likelihood, but
DistNet also restricts the output distribution to directly be a member of that
parametric distribution class. From Figures \ref{fig_low_samples} and
\ref{fig_censoring_samples}, we can see that the choice of the parametric
distribution used has a greater impact on the predictive performance on
DistNet, than it does for the Bayesian models.  It's almost certain that RTDs
do not perfectly match known parametric distributions.  From the data, we can
see that putting a restriction on the type of distributions the model can
predict can have a negative impact, as is the case for DistNet. 

We finally give some insight as to why we believe our
Bayesian models outperform DistNet, even when the same parametric distribution
is used. For low samples per instance and various levels of censoring, Figures
\ref{fig_low_samples} and \ref{fig_censoring_samples} respectively show where
the density of the predicted RTDs lie.  We plot the percentage of total area
under the RTDs outside the range $[0,T]$, where $T$ is 1.5 times the maximum
observed runtime for each instance.  We expect that it should be improbable to
see many runtimes outside this range, and thus predicted RTDs should give
little density outside this range if the density is to be informative.  When a
low number of samples is used, we can see from Figure \ref{fig_low_samples}
that in some cases, DistNet is unable to give an informative predictive RTD,
as a significant amount of density is outside the predictable range.  For the
censored cases, Figure \ref{fig_censoring_samples} shows similar results.

We summarize in that both DistNet and our Bayesian variant
perform similarly if given enough non-censored data. As the level of censoring
increases and as the number of samples per instance decreases, our Bayesian
model starts to outperform DistNet and better matches the empirical RTDs.  It
appears from our results that the discrepancy in performance between the
Bayesian and non-Bayesian versions is more affected by the number of samples
used.


\section{Conclusions and Future Work}

In this paper, we have shown that the existing state-of-the-art RTD prediction
model can be extended to the Bayesian setting.  Our Bayesian model
Bayes DistNet outperforms
the previous state-of-the-art in the low observation setting, as well as
with handling censored examples. The Bayesian model also lifts some of the
restrictions that occur when previous models could only output an explicit
parametric distribution, when many randomized algorithms do not perfectly
follow these parametric distributions.

Both DistNet and Bayes DistNet require a parametric distribution $\mathcal{F}$
to quantify the likelihood in the loss function. As previously
mentioned, we chose for our Bayesian network to output predicted runtimes
directly instead of $\mathcal{F}$'s parameters. We see as future work coming
up with an efficient method to have the network output $\mathcal{F}$'s
parameters, and compare it with our method.
We would also like to see further extensions of
this work which do not depend on a restrictive class of parametric
distributions. Finally, existing algorithm portfolio methods
can be extended to utilize the features of our Bayesian
network, such as having access to the uncertainty in predicting runtime
distributions.

\bibliography{references}

\begin{thebibliography}{30}
\providecommand{\natexlab}[1]{#1}
\providecommand{\url}[1]{\texttt{#1}}
\providecommand{\urlprefix}{URL }
\expandafter\ifx\csname urlstyle\endcsname\relax
  \providecommand{\doi}[1]{doi:\discretionary{}{}{}#1}\else
  \providecommand{\doi}{doi:\discretionary{}{}{}\begingroup
  \urlstyle{rm}\Url}\fi

\bibitem[{Amodei et~al.(2016)Amodei, Olah, Steinhardt, Christiano, Schulman,
  and Man{\'e}}]{amodei2016concrete}
Amodei, D.; Olah, C.; Steinhardt, J.; Christiano, P.; Schulman, J.; and
  Man{\'e}, D. 2016.
\newblock Concrete problems in AI safety.

\bibitem[{Blei, Kucukelbir, and McAuliffe(2017)}]{blei2017variational}
Blei, D.~M.; Kucukelbir, A.; and McAuliffe, J.~D. 2017.
\newblock Variational inference: A review for statisticians.
\newblock \emph{Journal of the American statistical Association} 112(518):
  859--877.

\bibitem[{Blundell et~al.(2015)Blundell, Cornebise, Kavukcuoglu, and
  Wierstra}]{blundell15}
Blundell, C.; Cornebise, J.; Kavukcuoglu, K.; and Wierstra, D. 2015.
\newblock Weight uncertainty in neural network.
\newblock In Bach, F.; and Blei, D., eds., \emph{Proceedings of the 32nd
  International Conference on Machine Learning}, volume~37 of \emph{Proceedings
  of Machine Learning Research}, 1613--1622. Lille, France: PMLR.
\newblock \urlprefix\url{http://proceedings.mlr.press/v37/blundell15.html}.

\bibitem[{Eggensperger, Lindauer, and Hutter(2018)}]{eggensperger2017neural}
Eggensperger, K.; Lindauer, M.; and Hutter, F. 2018.
\newblock Neural networks for predicting algorithm runtime distributions.
\newblock In \emph{Proceedings of the Twenty-Seventh International Joint
  Conference on Artificial Intelligence, {IJCAI-18}}, 1442--1448. International
  Joint Conferences on Artificial Intelligence Organization.
\newblock \doi{10.24963/ijcai.2018/200}.
\newblock \urlprefix\url{https://doi.org/10.24963/ijcai.2018/200}.

\bibitem[{Fortunato, Blundell, and Vinyals(2017)}]{fortunato2017bayesian}
Fortunato, M.; Blundell, C.; and Vinyals, O. 2017.
\newblock {B}ayesian recurrent neural networks.

\bibitem[{Gagliolo and Schmidhuber(2005)}]{gagliolo2005neural}
Gagliolo, M.; and Schmidhuber, J. 2005.
\newblock A neural network model for inter-problem adaptive online time
  allocation.
\newblock In \emph{International Conference on Artificial Neural Networks},
  7--12. Springer.

\bibitem[{Gagliolo and Schmidhuber(2006)}]{Gagliolo2006}
Gagliolo, M.; and Schmidhuber, J. 2006.
\newblock {Impact of censored sampling on the performance of restart
  strategies}.
\newblock \emph{Lecture Notes in Computer Science (including subseries Lecture
  Notes in Artificial Intelligence and Lecture Notes in Bioinformatics)} 4204
  LNCS: 167--181.
\newblock ISSN 16113349.
\newblock \doi{10.1007/11889205_14}.

\bibitem[{Gebser, Kaufmann, and Schaub(2012)}]{Gebser2012}
Gebser, M.; Kaufmann, B.; and Schaub, T. 2012.
\newblock Conflict-driven answer set solving: From theory to practice.
\newblock \emph{Artif. Intell.} 187–188: 52–89.
\newblock ISSN 0004-3702.
\newblock \doi{10.1016/j.artint.2012.04.001}.

\bibitem[{Gelfand and Smith(1990)}]{gelfand1990sampling}
Gelfand, A.~E.; and Smith, A.~F. 1990.
\newblock Sampling-based approaches to calculating marginal densities.
\newblock \emph{Journal of the American statistical association} 85(410):
  398--409.

\bibitem[{Gerevini and Serina(2002)}]{Gerevini2002}
Gerevini, A.; and Serina, I. 2002.
\newblock {LPG}: A planner based on local search for planning graphs with
  action costs.
\newblock In \emph{Proceedings of the Sixth International Conference on
  Artificial Intelligence Planning Systems}, AIPS’02, 13–22. AAAI Press.
\newblock ISBN 1577351428.

\bibitem[{Glorot, Bordes, and Bengio(2011)}]{glorot2011deep}
Glorot, X.; Bordes, A.; and Bengio, Y. 2011.
\newblock Deep sparse rectifier neural networks.
\newblock In \emph{Proceedings of the fourteenth international conference on
  artificial intelligence and statistics}, 315--323.

\bibitem[{Gomes, Selman, and Kautz(1998)}]{10.5555/295240.295710}
Gomes, C.~P.; Selman, B.; and Kautz, H. 1998.
\newblock Boosting combinatorial search through randomization.
\newblock In \emph{Proceedings of the Fifteenth National/Tenth Conference on
  Artificial Intelligence/Innovative Applications of Artificial Intelligence},
  AAAI ’98/IAAI ’98, 431–437. USA: American Association for Artificial
  Intelligence.
\newblock ISBN 0262510987.

\bibitem[{Graves(2011)}]{graves2011practical}
Graves, A. 2011.
\newblock Practical variational inference for neural networks.
\newblock In \emph{Advances in neural information processing systems},
  2348--2356.

\bibitem[{Harvey(1995)}]{Harvey1995NonsystematicBS}
Harvey, W.~D. 1995.
\newblock \emph{Nonsystematic backtracking search}.
\newblock Ph.D. thesis, Stanford, CA, USA.

\bibitem[{Hinton and Van~Camp(1993)}]{hinton1993keeping}
Hinton, G.~E.; and Van~Camp, D. 1993.
\newblock Keeping the neural networks simple by minimizing the description
  length of the weights.
\newblock In \emph{Proceedings of the sixth annual conference on Computational
  learning theory}, 5--13.

\bibitem[{Ioffe and Szegedy(2015)}]{ioffe2015batch}
Ioffe, S.; and Szegedy, C. 2015.
\newblock Batch normalization: Accelerating deep network training by reducing
  internal covariate shift.

\bibitem[{Kendall and Gal(2017)}]{kendall2017uncertainties}
Kendall, A.; and Gal, Y. 2017.
\newblock What uncertainties do we need in {B}ayesian deep learning for
  computer vision?
\newblock In \emph{Advances in neural information processing systems},
  5574--5584.

\bibitem[{Kingma and Welling(2014)}]{Kingma2014AutoEncodingVB}
Kingma, D.~P.; and Welling, M. 2014.
\newblock Auto-Encoding Variational Bayes.
\newblock \emph{CoRR} abs/1312.6114.

\bibitem[{Klein and Moeschberger(2003)}]{klein03}
Klein, J.~P.; and Moeschberger, M.~L. 2003.
\newblock \emph{Survival analysis techniques for censored and truncated data}.
\newblock Second edition.

\bibitem[{Luby, Sinclair, and Zuckerman(1993)}]{Luby1993}
Luby, M.; Sinclair, A.; and Zuckerman, D. 1993.
\newblock {Optimal speedup of Las Vegas algorithms}.
\newblock \emph{Information Processing Letters} 47(4): 173--180.
\newblock ISSN 00200190.
\newblock \doi{10.1016/0020-0190(93)90029-9}.

\bibitem[{MacKay(1992)}]{mackay1992practical}
MacKay, D.~J. 1992.
\newblock A practical {B}ayesian framework for backpropagation networks.
\newblock \emph{Neural computation} 4(3): 448--472.

\bibitem[{Neal(2012)}]{neal2012bayesian}
Neal, R.~M. 2012.
\newblock \emph{{B}ayesian learning for neural networks}, volume 118.
\newblock Springer Science \& Business Media.

\bibitem[{Nikoli{\'{c}}, Mari{\'{c}}, and
  Jani{\v{c}}i{\'{c}}(2009)}]{10.1007/978-3-642-02777-2_31}
Nikoli{\'{c}}, M.; Mari{\'{c}}, F.; and Jani{\v{c}}i{\'{c}}, P. 2009.
\newblock Instance-based selection of policies for {SAT} solvers.
\newblock In Kullmann, O., ed., \emph{Theory and Applications of Satisfiability
  Testing - SAT 2009}, 326--340. Berlin, Heidelberg: Springer Berlin
  Heidelberg.
\newblock ISBN 978-3-642-02777-2.

\bibitem[{Opper and Archambeau(2009)}]{opper2009variational}
Opper, M.; and Archambeau, C. 2009.
\newblock The variational Gaussian approximation revisited.
\newblock \emph{Neural computation} 21(3): 786--792.

\bibitem[{Paszke et~al.(2019)Paszke, Gross, Massa, Lerer, Bradbury, Chanan,
  Killeen, Lin, Gimelshein, Antiga, Desmaison, Kopf, Yang, DeVito, Raison,
  Tejani, Chilamkurthy, Steiner, Fang, Bai, and Chintala}]{NEURIPS2019_9015}
Paszke, A.; Gross, S.; Massa, F.; Lerer, A.; Bradbury, J.; Chanan, G.; Killeen,
  T.; Lin, Z.; Gimelshein, N.; Antiga, L.; Desmaison, A.; Kopf, A.; Yang, E.;
  DeVito, Z.; Raison, M.; Tejani, A.; Chilamkurthy, S.; Steiner, B.; Fang, L.;
  Bai, J.; and Chintala, S. 2019.
\newblock PyTorch: An imperative style, high-performance deep learning library.
\newblock In \emph{Advances in Neural Information Processing Systems 32},
  8024--8035. Curran Associates, Inc.
\newblock
  \urlprefix\url{http://papers.neurips.cc/paper/9015-pytorch-an-imperative-style-high-performance-deep-learning-library.pdf}.

\bibitem[{Penberthy and Weld(1994)}]{Penberthy1994}
Penberthy, J.~S.; and Weld, D.~S. 1994.
\newblock Temporal planning with continuous change.
\newblock In \emph{Proceedings of the Twelfth AAAI National Conference on
  Artificial Intelligence}, AAAI’94, 1010–1015. AAAI Press.

\bibitem[{Prechelt(1998)}]{prechelt1998automatic}
Prechelt, L. 1998.
\newblock Automatic early stopping using cross validation: quantifying the
  criteria.
\newblock \emph{Neural Networks} 11(4): 761--767.

\bibitem[{Shridhar, Laumann, and Liwicki(2019)}]{shridhar2019comprehensive}
Shridhar, K.; Laumann, F.; and Liwicki, M. 2019.
\newblock A comprehensive guide to {B}ayesian convolutional neural network with
  variational inference.

\bibitem[{Srivastava et~al.(2014)Srivastava, Hinton, Krizhevsky, Sutskever, and
  Salakhutdinov}]{srivastava2014dropout}
Srivastava, N.; Hinton, G.; Krizhevsky, A.; Sutskever, I.; and Salakhutdinov,
  R. 2014.
\newblock Dropout: a simple way to prevent neural networks from overfitting.
\newblock \emph{The journal of machine learning research} 15(1): 1929--1958.

\bibitem[{Xu et~al.(2008)Xu, Hutter, Hoos, and Leyton-Brown}]{xu2008satzilla}
Xu, L.; Hutter, F.; Hoos, H.~H.; and Leyton-Brown, K. 2008.
\newblock {SAT}zilla: portfolio-based algorithm selection for SAT.
\newblock \emph{Journal of artificial intelligence research} 32: 565--606.

\end{thebibliography}

\clearpage
\onecolumn
\section*{Appendix}

\begin{table}[h]
\centering
\resizebox{\textwidth}{!}{
\begin{tabular}{llllLLLL}
    \toprule
    \textbf{Scenario} & \textbf{Num Samples} & \textbf{Model} & \textbf{Distribution} &  \textbf{NLLH} &  \textbf{KLD} & \textbf{D-KS} & \textbf{Mass} \\
    \midrule
clasp\_factoring & 1  & Bayes DistNet & InverseGaussian &   0.587,0.089 &  0.590,0.060 &  0.495,0.035 &  0.284,0.030 \\
         &    &         & Lognormal &   0.459,0.123 &  0.508,0.081 &  0.450,0.043 &  0.226,0.047 \\
         &    & DistNet & InverseGaussian &   0.737,0.205 &  0.770,0.139 &  0.521,0.042 &  0.295,0.044 \\
         &    &         & Lognormal &   1.236,0.557 &  1.061,0.383 &  0.590,0.051 &  0.403,0.058 \\ \midrule
clasp\_factoring & 2  & Bayes DistNet & InverseGaussian &   0.316,0.180 &  0.419,0.115 &  0.391,0.057 &  0.171,0.073 \\
         &    &         & Lognormal &   0.113,0.165 &  0.295,0.103 &  0.326,0.050 &  0.092,0.062 \\
         &    & DistNet & InverseGaussian &   0.436,0.152 &  0.565,0.106 &  0.454,0.038 &  0.231,0.029 \\
         &    &         & Lognormal &   0.705,0.137 &  0.691,0.090 &  0.523,0.041 &  0.348,0.044 \\ \midrule
clasp\_factoring & 4  & Bayes DistNet & InverseGaussian &  -0.040,0.060 &  0.203,0.037 &  0.277,0.021 &  0.037,0.018 \\
         &    &         & Lognormal &  -0.054,0.051 &  0.196,0.032 &  0.274,0.015 &  0.032,0.015 \\
         &    & DistNet & InverseGaussian &   0.224,0.065 &  0.412,0.044 &  0.392,0.026 &  0.180,0.022 \\
         &    &         & Lognormal &   0.289,0.213 &  0.431,0.133 &  0.414,0.069 &  0.198,0.086 \\ \midrule
clasp\_factoring & 8  & Bayes DistNet & InverseGaussian &  -0.110,0.021 &  0.160,0.016 &  0.251,0.012 &  0.021,0.004 \\
         &    &         & Lognormal &  -0.112,0.025 &  0.160,0.017 &  0.253,0.012 &  0.019,0.006 \\
         &    & DistNet & InverseGaussian &  -0.001,0.202 &  0.275,0.130 &  0.293,0.069 &  0.077,0.084 \\
         &    &         & Lognormal &  -0.135,0.022 &  0.192,0.018 &  0.254,0.012 &  0.014,0.003 \\ \midrule
clasp\_factoring & 16 & Bayes DistNet & InverseGaussian &  -0.127,0.023 &  0.149,0.016 &  0.245,0.014 &  0.018,0.004 \\
         &    &         & Lognormal &  -0.136,0.022 &  0.145,0.015 &  0.245,0.012 &  0.014,0.004 \\
         &    & DistNet & InverseGaussian &  -0.140,0.096 &  0.207,0.054 &  0.255,0.035 &  0.019,0.044 \\
         &    &         & Lognormal &  -0.156,0.022 &  0.179,0.018 &  0.248,0.013 &  0.010,0.002 \\ \midrule
lpg-zeno & 1  & Bayes DistNet & InverseGaussian &   0.562,0.167 &  0.923,0.126 &  0.659,0.060 &  0.174,0.065 \\
         &    &         & Lognormal &   0.520,0.292 &  0.901,0.208 &  0.661,0.066 &  0.126,0.023 \\
         &    & DistNet & InverseGaussian &   1.779,1.106 &  1.835,0.762 &  0.790,0.073 &  0.438,0.087 \\
         &    &         & Lognormal &   1.106,0.368 &  1.290,0.253 &  0.738,0.065 &  0.488,0.078 \\ \midrule
lpg-zeno & 2  & Bayes DistNet & InverseGaussian &   0.407,0.316 &  0.827,0.220 &  0.636,0.079 &  0.107,0.010 \\
         &    &         & Lognormal &   0.391,0.369 &  0.823,0.253 &  0.635,0.082 &  0.085,0.014 \\
         &    & DistNet & InverseGaussian &   0.970,0.700 &  1.281,0.498 &  0.723,0.088 &  0.333,0.058 \\
         &    &         & Lognormal &   0.868,0.403 &  1.154,0.276 &  0.714,0.070 &  0.378,0.060 \\ \midrule
lpg-zeno & 4  & Bayes DistNet & InverseGaussian &  -0.009,0.234 &  0.605,0.150 &  0.572,0.049 &  0.050,0.026 \\
         &    &         & Lognormal &  -0.021,0.422 &  0.612,0.257 &  0.565,0.060 &  0.043,0.135 \\
         &    & DistNet & InverseGaussian &   0.126,0.136 &  0.702,0.101 &  0.596,0.035 &  0.219,0.026 \\
         &    &         & Lognormal &   0.126,0.303 &  0.716,0.193 &  0.591,0.057 &  0.143,0.106 \\ \midrule
lpg-zeno & 8  & Bayes DistNet & InverseGaussian &  -0.627,0.183 &  0.217,0.118 &  0.319,0.139 &  0.003,0.002 \\
         &    &         & Lognormal &  -0.611,0.192 &  0.232,0.124 &  0.321,0.145 &  0.002,0.002 \\
         &    & DistNet & InverseGaussian &  -0.265,0.093 &  0.438,0.079 &  0.429,0.074 &  0.139,0.019 \\
         &    &         & Lognormal &  -0.613,0.239 &  0.266,0.159 &  0.327,0.137 &  0.013,0.027 \\ \midrule
lpg-zeno & 16 & Bayes DistNet & InverseGaussian &  -0.672,0.119 &  0.186,0.075 &  0.298,0.117 &  0.001,0.001 \\
         &    &         & Lognormal &  -0.664,0.127 &  0.197,0.080 &  0.297,0.124 &  0.001,0.001 \\
         &    & DistNet & InverseGaussian &  -0.446,0.120 &  0.350,0.084 &  0.374,0.079 &  0.073,0.043 \\
         &    &         & Lognormal &  -0.682,0.160 &  0.234,0.120 &  0.304,0.119 &  0.000,0.000 \\
\bottomrule
\end{tabular}
}
\caption{Table corresponding to Figure \ref{fig_low_samples}.
Comparison of DistNet and Bayes DistNet, for various
numbers of observed runtimes per instance. Metrics are averaged across
10-folds, repeated for multiple seeds. From left to right:
NLLH, KL-Divergence, KS-Distance, and percentage of density area outside the
expected range of 0 to 1.5 times the maximum observed runtime for each
instance.}
\label{table_num_samples}
\end{table}

\begin{table}[h]
\centering
\resizebox{\textwidth}{!}{
    \begin{tabular}{llllLLLL}
    \toprule
    \textbf{Scenario} & \textbf{Lower Bound} & \textbf{Model} & \textbf{Distribution} &  \textbf{NLLH} &  \textbf{KLD} & \textbf{D-KS} & \textbf{Mass} \\
    \midrule
clasp\_factoring & 0  & Bayes DistNet & InverseGaussian &  -0.110,0.021 &  0.160,0.016 &  0.251,0.012 &  0.021,0.004 \\
         &    &         & Lognormal &  -0.112,0.025 &  0.160,0.017 &  0.253,0.012 &  0.019,0.006 \\
         &    & DistNet & InverseGaussian &  -0.001,0.202 &  0.275,0.130 &  0.293,0.069 &  0.077,0.084 \\
         &    &         & Lognormal &  -0.135,0.022 &  0.192,0.018 &  0.254,0.012 &  0.014,0.003 \\ \midrule
clasp\_factoring & 20 & Bayes DistNet & InverseGaussian &  -0.078,0.025 &  0.175,0.017 &  0.257,0.012 &  0.033,0.006 \\
         &    &         & Lognormal &  -0.095,0.022 &  0.166,0.016 &  0.254,0.012 &  0.026,0.005 \\
         &    & DistNet & InverseGaussian &   0.087,0.150 &  0.315,0.099 &  0.333,0.058 &  0.125,0.064 \\
         &    &         & Lognormal &  -0.103,0.036 &  0.196,0.020 &  0.264,0.017 &  0.031,0.015 \\ \midrule
clasp\_factoring & 40 & Bayes DistNet & InverseGaussian &  -0.055,0.033 &  0.188,0.021 &  0.264,0.016 &  0.041,0.010 \\
         &    &         & Lognormal &  -0.087,0.025 &  0.172,0.017 &  0.257,0.012 &  0.029,0.007 \\
         &    & DistNet & InverseGaussian &   0.167,0.065 &  0.361,0.040 &  0.372,0.031 &  0.170,0.031 \\
         &    &         & Lognormal &   0.058,0.179 &  0.285,0.102 &  0.329,0.072 &  0.105,0.086 \\ \midrule
clasp\_factoring & 60 & Bayes DistNet & InverseGaussian &   0.059,0.037 &  0.254,0.022 &  0.302,0.014 &  0.084,0.014 \\
         &    &         & Lognormal &  -0.037,0.037 &  0.200,0.022 &  0.275,0.016 &  0.047,0.013 \\
         &    & DistNet & InverseGaussian &   0.227,0.045 &  0.399,0.030 &  0.409,0.019 &  0.207,0.019 \\
         &    &         & Lognormal &   0.395,0.221 &  0.487,0.134 &  0.455,0.072 &  0.259,0.091 \\ \midrule
clasp\_factoring & 80 & Bayes DistNet & InverseGaussian &   0.372,0.050 &  0.451,0.033 &  0.418,0.014 &  0.205,0.020 \\
         &    &         & Lognormal &   0.285,0.057 &  0.397,0.036 &  0.397,0.018 &  0.164,0.021 \\
         &    & DistNet & InverseGaussian &   0.345,0.072 &  0.488,0.051 &  0.459,0.029 &  0.253,0.028 \\
         &    &         & Lognormal &   0.663,0.138 &  0.661,0.090 &  0.545,0.038 &  0.358,0.045 \\ \midrule
lpg-zeno & 0  & Bayes DistNet & InverseGaussian &  -0.627,0.183 &  0.217,0.118 &  0.319,0.139 &  0.003,0.002 \\
         &    &         & Lognormal &  -0.611,0.192 &  0.232,0.124 &  0.321,0.145 &  0.002,0.002 \\
         &    & DistNet & InverseGaussian &  -0.265,0.093 &  0.438,0.079 &  0.429,0.074 &  0.139,0.019 \\
         &    &         & Lognormal &  -0.613,0.239 &  0.266,0.159 &  0.327,0.137 &  0.013,0.027 \\ \midrule
lpg-zeno & 20 & Bayes DistNet & InverseGaussian &  -0.509,0.217 &  0.300,0.141 &  0.347,0.135 &  0.019,0.020 \\
         &    &         & Lognormal &  -0.561,0.226 &  0.284,0.138 &  0.328,0.153 &  0.007,0.009 \\
         &    & DistNet & InverseGaussian &  -0.158,0.175 &  0.493,0.122 &  0.465,0.093 &  0.170,0.041 \\
         &    &         & Lognormal &  -0.505,0.338 &  0.352,0.186 &  0.343,0.162 &  0.038,0.074 \\ \midrule
lpg-zeno & 40 & Bayes DistNet & InverseGaussian &  -0.155,0.235 &  0.501,0.158 &  0.446,0.121 &  0.109,0.033 \\
         &    &         & Lognormal &  -0.214,0.281 &  0.483,0.184 &  0.432,0.134 &  0.086,0.032 \\
         &    & DistNet & InverseGaussian &  -0.000,0.226 &  0.579,0.155 &  0.525,0.089 &  0.226,0.053 \\
         &    &         & Lognormal &   0.139,0.393 &  0.677,0.244 &  0.540,0.118 &  0.250,0.117 \\ \midrule
lpg-zeno & 60 & Bayes DistNet & InverseGaussian &   0.234,0.311 &  0.740,0.207 &  0.541,0.119 &  0.192,0.042 \\
         &    &         & Lognormal &   0.201,0.341 &  0.729,0.225 &  0.529,0.126 &  0.181,0.045 \\
         &    & DistNet & InverseGaussian &   0.194,0.204 &  0.714,0.141 &  0.596,0.073 &  0.286,0.046 \\
         &    &         & Lognormal &   0.487,0.287 &  0.897,0.186 &  0.630,0.084 &  0.359,0.068 \\ \midrule
lpg-zeno & 80 & Bayes DistNet & InverseGaussian &   0.993,0.421 &  1.214,0.277 &  0.684,0.088 &  0.315,0.054 \\
         &    &         & Lognormal &   0.926,0.391 &  1.179,0.259 &  0.677,0.092 &  0.290,0.049 \\
         &    & DistNet & InverseGaussian &   2.663,1.033 &  2.419,0.698 &  0.774,0.060 &  0.507,0.081 \\
         &    &         & Lognormal &   1.330,0.394 &  1.449,0.260 &  0.757,0.066 &  0.497,0.074 \\
\bottomrule
\end{tabular}
}
\caption{Table corresponding to Figure \ref{fig_censoring_samples}.
Comparison of DistNet and Bayes DistNet, for various
levels of censoring, each with 8 observations per instance.
Metrics are averaged across
10-folds, repeated for multiple seeds. From left to right:
NLLH, KL-Divergence, KS-Distance, and percentage of density area outside the
expected range of 0 to 1.5 times the maximum observed runtime for each
instance.}
\label{table_censored}
\end{table}

\end{document}